**Title**

- Identifying social isolation themes in NVDRS text narratives using topic modeling and text-classification methods
- Identifying Social Isolation in NVDRS with NLP

**Authors**

Drew Walker [1*], Swati Rajwal [2], Sudeshna Das [1], Snigdha Peddireddy [3], Abeed Sarker [1,4]

Corresponding Author: Drew Walker, PhD. E-mail: Andrew.walker@emory.edu

**Affiliations**

1. Department of Biomedical Informatics, School of Medicine, Emory University, Atlanta GA, USA
2. Department of Computer Science, Emory College of Arts and Sciences, Emory University, Atlanta, GA, USA
3. Department of Behavioral, Social, Health Education Sciences, Rollins School of Public Health, Emory University, Atlanta GA, USA
4. Department of Biomedical Engineering, Georgia Institute of Technology and Emory University, Atlanta, GA, USA

**Abstract**

Social isolation and loneliness, which have been increasing in recent years strongly contribute toward suicide rates. Although social isolation and loneliness are not currently recorded within the US National Violent Death Reporting System's (NVDRS) structured variables, natural language processing (NLP) techniques can be used to identify these constructs in law enforcement and coroner/medical examiner narratives. Using topic modeling, lexicon development and supervised learning classifiers, we developed high-quality classifiers (average F1: .86, accuracy: .82). Evaluating over 300,000 suicides from 2002 to 2020, we identified 1,198 mentioning chronic social isolation. Decedents had higher odds of chronic social isolation classification if they were men (OR=1.44; CI: 1.24, 1.69, p<.0001), gay (OR=3.68; 1.97, 6.33, p<.0001), or were divorced (OR=3.34; 2.68, 4.19, p<.0001). We found significant predictors for other social isolation topics of recent or impending divorce, child custody loss, eviction or recent move, and break-up. Our methods can improve surveillance and prevention of social isolation and loneliness in the United States.

**Teaser**

Natural language processing identified trends and predictors of social isolation and loneliness among US suicide records.

# MAIN TEXT

## INTRODUCTION

Social isolation is a critical factor that contributes to the United States' mental health and suicide crisis. In 2023, the US Surgeon General released an advisory highlighting the epidemic of loneliness and social isolation and proposing a call to action for intervention frameworks and policy priorities.[1] He emphasized the Interpersonal Theory of Suicide's assertion that social isolation is the most important contributing factor to suicide across all age and demographic groups.[1,2] Social isolation and its intense, well-documented negative effects, may be experienced more acutely during specific critical life events, and these experiences may vary by race, gender, or other demographic factors.[3,4] Despite mounting evidence on the detrimental effects of social isolation, identification and measurement of this phenomenon, and its impacts on mortality and morbidity at an epidemiological scale are difficult to measure, given existing survey methods.[5,6] Recent reviews have called for public health researchers to pursue novel methods to understand evolving trends in how social isolation contributes to suicide and mental health, and how these trends differ across populations and demographic factors.[5,7,8]

The National Violent Death Reporting System (NVDRS) has historically been one of the most comprehensive US data sources on suicide and violent deaths.[9] NVDRS pools over 600 data elements, abstracted from coroner medical examiner and law enforcement narratives, toxicology reports, and death certificates. Current research on NVDRS has used the dataset to explore contexts surrounding suicide, in an attempt to understand common circumstances and co-occurring alcohol and substance use.[10] NVDRS collects data on demographics and a wide variety of related circumstances, ranging from "relationship problems; mental health conditions and treatment; toxicology results; and life stressors, including recent money- or work-related or physical health problems."[11] Still, there are limits to structured data, including information about circumstances related to social isolation and related events.[10,12] The NVDRS contains extensive text data from law enforcement (LE) and coroner medical examiner (CME) incident narratives, including information on circumstances, other suspects or individuals on scene, and additional details not captured in structured data. These text fields may vary in the level of detail but often provide additional information not captured within structured variables. Despite the potential insights from LE/CME narratives, there is significant cost in time and effort required by abstractors to manually review a large range of concepts within this volume of highly variable and lengthy text data.[12]Advances in natural language processing have enabled researchers in recent years to apply computational methods to mine information from narratives in NVDRS, illuminating important trends related to mental health, violent death, and suicide.[12–14]

The objective of this study was to explore circumstances related to the intersection of social isolation and suicide. To achieve this objective, we developed and applied natural language processing techniques to identify social isolation-related circumstances and events from free-text LE/CME narratives, recorded within the NVDRS for suicide decedents identified from 2002 to 2020. After development of identification methods, we applied models to assess longitudinal trends and associations with relevant demographic factors to identify at-risk groups for social isolation-related suicide.

**RESULTS**

Table 1 provides descriptive statistics for the 306,817 decedents who died by suicide recorded in the NVDRS dataset from 2002 to 2020. Most decedents in this sample were male (78.1%), with an average age of 46.3 years, and largely were classified as White (82.1%), Black (6.5%), or Hispanic (6.3%) decedents.

***BERTopic Topic Modeling***

BERTopic topic modeling generated 64 topics within the highest performance coherence model, which was trained upon the shorter text fields of LE/CME suicide circumstance summaries. LE circumstance summaries (n = 20,731, 6.8% of decedents) and CME circumstance summaries (n=30,584, 10.0%) provide shorter text summarizations (40-50 characters) of relevant contexts related to the suicides mentioned within the corresponding full LE/CME narratives. The shorter text within these fields allowed for more easily understood and coherent topic modeling results, which could then be identified by subject matter experts, and used to construct lexicons of words and phrases to search across a larger sample of full-length LE/CME narratives.

Shorter text circumstance summary fields were used for topic modeling due to the high amount of noise assessed in longer-length full LE/CME narratives. After reviewing topics, we arrived at six distinct narrative topics to apply towards classifying the full LE/CME narratives related to social isolation: 1) chronic social isolation, 2) recent or impending divorce, 3) eviction or move, 4) break-up, 5) child custody loss, and 6) pet loss. These classifier labels were chosen due to their relevance to social isolation and value added to identifying phenomena not currently captured by existing NVDRS structured variables.

### Regex search and annotation process

Table 2 displays results of regular expression search matches, along with the percentage of the sample which were annotated as related to each topic, interrater reliability for the first 50 samples, and an example sentence (edited to provide anonymity) from positively classified narrative matches for each topic. When matching with the regular expression lexicons, we had the highest frequencies of matches with 1) recent eviction or move, (N = 29,913 decedent narrative matches; 9.76% of sample narratives), 2) recent or impending divorce or separation (15,263; 4.98%), and 3) recent break-up (12,279; 4.0%). Lower frequencies of matches were found among the topics of child custody loss (5,322, 1.73% of sample), recent pet loss (1,355, 0.44%), and chronic social isolation (1,198, 0.39%).

Following manual annotation to assess relevance to topics of interest, nearly all 100 samples with break-up/social isolation were found to be relevant to the topic, while divorce (66%), child custody loss (50%), eviction/move (36%), and loss of pet (28%) resulted in fewer narratives that actually contained references to the topic of interest. Interrater annotator agreement was relatively high across all topics, ranging from 72 to 98%, (kappa range: .24, .85).

### Sample Annotation Results

Supplemental Table 1 describes social isolation and related event classifier performance results, following the annotation of 100 samples of each of the 6 topics. Overall, across all models, best results were achieved from logistic regression and naive bayes models. Of all the topics, we achieved highest performance results in social isolation, child custody loss, break-up, and divorce, with macro F1 performance ranging from .87 to 1, accuracy from .66 to 1, recall from .75 to 1, and positive-class precision from .75 to 1. Recent eviction or move performed slightly worse, with the pet loss classifier performing the worst, only predicting negative classes. For this reason, the pet loss topic was dropped from the predictive analyses.

### Predictions, Distributions on entire Dataset

Supplemental Table 2 describes the frequencies of regular expression matches and refined classifier predictions across the entire dataset, normalized per rate of 1000 suicides in the overall dataset. The most frequently predicted topics included 1) recent or impending divorce (50.0 per 1000 decedents), 2) recent breakup (40.1 per 1000 decedents), and recent eviction or move (30.9 per 1000 decedents). The least frequently predicted classes were child custody loss (4.0 per 1000 decedents), and chronic social isolation (3.9 per 1000 decedents).

Figure 1 displays trends in suicide events by topic from 2002 to 2020. Over this time, across all topics, we observed a gradual increase in rates over time, with several peaks in topic classifications per year. Recent breakup, recent/impending divorce, and recent eviction/move were the most frequently observed topics over time, despite decreases by 2020. We observed peaks in suicide narratives mentioning recent/impending divorces in 2010; and recent evictions/moves in 2016 through 2017. Chronic social isolation and child custody loss peaked in 2020 and 2019, respectively, though both topics had relatively much lower frequencies than others.

### Demographic predictors of social isolation-related event narrative classification

Decedents flagged with social isolation in their LE/CME narrative, on average, were 1.28 years younger (95% CI: -0.23, -2.32, p = .016) than those without. For divorce, decedents were on average 1.44 years younger (95% CI: -1.74, -1.14, p<.0001). Eviction classifications were associated with being younger (95% CI: -1.43, -.67, p<.0001). Larger differences in age were found among decedents with child custody loss classifications, where decedents were 9.28 years younger on average (95% CI: -10.31, -8.25, p<.0001), and those with breakup classifications, where decedents were on average 15.11 years younger (95% CI: -15.44, -14.79, p<.0001). We next examined non-continuous demographic variables and whether the decedent had a social isolation-classified LE/CME narrative (Table 3).

### Chronic Social Isolation

Men had higher odds of chronic social isolation classification (Odds Ratio (OR)=1.44, 95% CI: 1.24, 1.69, p<.0001); decedents identified as gay had higher odds (OR=3.68, 95% CI: 1.97, 6.33, p<.0001) compared with heterosexual decedents; and those who were divorced had higher odds (OR=3.34, 95% CI: 2.68, 4.19, p<.0001) compared with decedents who were married. The odds of having chronic social isolation classification was higher for married but separated decedents (OR=3.43, 95% CI: 2.24, 5.09, p<.0001), for widowed decedents (OR=3.04, 95% CI: 2.21, 4.14, p<.0001), for never married decedents (OR=5.81, 95% CI: 4.78, 7.12, p<.0001), and for decedents identified as not currently in a relationship (OR=6.97, 95% CI: 5.61, 8.69, p<.0001). Decedents identified as non-Hispanic Black or African American had lower odds of chronic social isolation classification (OR=0.62, 95% CI: .46, .81, p=.001) compared with non-Hispanic White decedents.

### Recent or Impending Divorce

Compared to married decedents, those identified as married but separated had higher odds of having recent or impending divorce narrative classification (OR=6.61, 95% CI: 6.27, 6.96, p<.0001); compared with decedents in a relationship, those not currently in a relationship had higher odds of having the recent or impending divorce classification (OR=1.79, 95%CI: 1.69, 1.89, p<.0001); men had higher odds of having recent or impending divorce classification than women (OR=1.31, 95% CI: 1.26, 1.37, p<.0001).

Additionally, several factors showed decreased odds of impending or recent divorce classification. Compared to decedents identified as White, most other racial groups had reduced odds of having recent or impending divorce classification. Decedents identified as gay had lower odds of having recent or impending divorce classification (OR=0.28, 95% CI: .19, .41, p<.0001) compared to heterosexual decedents; decedents identified as transgender had lower odds of having recent or impending divorce classification (OR=0.24, 95% CI: .10, .48, p<.0001). Decedents who were identified as divorced had lower odds of having recent or impending divorce narrative classification (OR=0.68, 95% CI: .65, .71, p<.0001); never married, single, or widowed all much lower odds of having recent or impending divorce classification (OR range: .04, .10, all p<.0001). Decedents identified as homeless had lower odds of recent or impending divorce narrative classification (OR=0.73, 95% CI: .61, .87, p<.0001); decedents identified with a physical health problem had lower odds of recent or impending divorce narrative classification (OR=0.54, 95% CI: .51, .57, p<.0001).

### Eviction or recent move

Decedents identified as Hispanic had higher odds of eviction classification (OR=1.18, 95% CI: 1.09, 1.28, p<.0001) compared to non-Hispanic White decedents. Decedents with marital statuses including divorce had higher odds of eviction or recent move (OR=1.33, 95% CI: 1.25, 1.40, p<.0001); decedents identified as married but separated had higher odds of eviction classification compared to married decedents (OR=2.34, 95% CI: 2.02, 2.48, p<.0001). Decedents who were not currently in a relationship had higher odds of eviction or recent move classification, compared to those currently in a relationship (OR=1.58, 95% CI: 1.47, 1.70, p <.0001). Decedents identified as homeless had higher odds of eviction or recent move narrative classification (OR=1.42, 95% CI: 1.20, 1.66, p<.0001). Finally, men had lower odds of having eviction or recent move classification compared to women (OR=0.88, 95%: .84, .92, p<.0001).

### *Break-up*

Men had higher odds of having breakup classification in narratives than women (OR=1.13, 95% CI: 1.08, 1.18, p<.0001).  Decedents identified as American Indian or Alaska Native, non-Hispanic had higher odds of having a break-up narrative classification compared to decedents identified as White (OR=1.46, 95% CI: 1.27, 1.68, p<.0001); decedents identified as Hispanic had higher odds of having break-up narrative classification compared to decedents identified as White (OR=1.61, 95% CI: 1.51, 1.72, p<.0001); decedents identified as two or more races, non-Hispanic had higher odds of having break-up narrative classifications compared to decedents identified as White (OR=1.74, 95% CI: 1.52, 2.00, p<.0001). Sexual orientation was also a significant predictor of break-up classification, where decedents identified as gay had higher odds of having a break-up narrative classification compared to heterosexual decedents (OR=1.90, 95% CI: 1.55, 2.31, p<.0001); decedents identified as lesbian had higher odds of having a break-up narrative classification compared to heterosexual decedents (OR=3.26, 95% CI: 2.57, 4.08, p<.0001). Across marital status, all decedents that were not indicated as married had increased odds, further detailed in Table 3. Not currently being in a relationship was associated with higher odds of having break-up narrative classification (OR=11.42, 95% CI: 10.79 12.08, p<.0001). Out of all demographic predictors, only having a physical health problem predicted reduced odds of break-up classification (OR=0.27, 95% CI: .25, .29, p<.0001).

### *Child custody loss*

Compared with decedents who were identified White, decedents identified as American Indian/Alaska Native, non-Hispanic had higher odds of having a child custody loss classification (OR=3.02, 95% CI: 2.20, 4.04, p<.0001); decedents who were identified as Hispanic had higher odds of having a child custody loss classification (OR=1.68, 95% CI: 1.38, 2.02, p<.0001). Marital status also predicted classification of child custody loss within narratives. Compared to those married, divorced marital status was associated with higher odds of child custody loss classification, (OR=2.06, 95% CI: 1.75, 2.42, p<.0001); married but separated was associated with higher odds of child custody loss classification (OR=2.65, 95% CI: 1.94, 3.53, p<.0001); never married was associated with higher odds of child custody loss classification (OR=1.75, 95% CI: 1.51, 2.03, p<.0001). Homelessness was associated with higher odds of having child custody loss classification, compared with decedents with housing (OR=2.08, 95% CI: 1.41, 2.94, p<.0001). Number of substances was associated with increased odds of having child custody loss classification, for each additional substance identified on scene (OR= 1.03, 95% CI: 1.01, 1.04, p<.0001). Sex identified in NVDRS was predictive of child custody loss narrative classification, where men had reduced odds compared with women (OR=0.57, 95% CI: .51, .65, p<.0001). Finally, decedents with physical health problems also had reduced odds of child custody loss classifications compared with decedents without physical health problems (OR=0.51, 95% CI: .43, .61, p<.0001).

## DISCUSSION

This study used NLP topic modeling to develop supervised classifiers of circumstances related to social isolation among LE/CME narratives for suicide decedents. Logistic regressions of classifiers applied to narratives revealed several demographic predictors with important associations that could be important to suicide surveillance and prevention efforts. Our NLP classifiers demonstrated validity with existing NVDRS variables and can be utilized by public health investigators to better examine predictors of social isolation that were difficult to identify via existing NVDRS variables alone. Our methodology provides a general framework for identifying other topics associated with suicide narratives and can better inform prevention efforts.

Logistic regressions of narrative topics with demographic predictors revealed several associations that may be important to suicide surveillance efforts. Our findings that male decedents had higher rates of social isolation narrative classification is in line with previous research suggesting that men experience higher rates of loneliness overall.[15] This effect was even higher for decedents identified as gay. This may be due to a variety of societal and systematic factors in which gay decedents may feel ostracized or stigmatized by their immediate social surroundings.[16] Research suggests that for many LGBT decedents, the spaces that seek to offer social support and connection, such as sports clubs and religious community organizations, may instead offer the opposite, with decedents experiencing rejection and stigma.[17] Results showing lack of/fractured intimate relationships are significant predictors of chronic social isolation are unsurprising, and support the validity of the classifier.

Our classifier for impending or recent divorce can likely pick up on a more nuanced relationship status than existing structural NVDRS variables. Results show that divorced decedents are less likely to have this classification, which suggests that the current "divorced" status variable may be capturing divorce that occurred in the more distant past relative to suicide. Instead, our impending/recent divorce classification is more closely related to the "Married, Civil/Domestic Partnership but separated" existing variable, and may be able to detect more recent changes. This is important to note, given the higher impact that recent divorces have been documented to have on suicides than more temporally distal divorces.[18] All non-White racial groups had lower odds of impending or recent divorce within suicide narratives, which falls in line with research suggesting that these groups have overall lower suicide risk following divorce, particularly among decedents with larger family support networks.[19] Reduced odds of classification of divorce among decedents identified as gay may also be related to same sex marriage not being broadly legal in the US until 2015. Narrative classifiers indicating recent break-ups showed high associations with single and not in relationship statuses. Among other demographic factors, we found greater likelihood of break-up classifications among gay and lesbian decedents. Within LGBT communities, break-ups may also result in more significant ruptures to other social community bonds than heterosexual decedents, causing the feelings of social isolation post-breakup to be more acute or severe.[20]

Despite being a rare narrative classification, child custody loss was highest among American Indians/Alaska Native persons and is particularly troubling. Among many other factors, this categorization appearing in suicide LE/CME narratives may be related to recent nationwide challenges surrounding navigating the intersecting jurisdictions of the Adoption and Safe Families Act and Indian Child Welfare Act, where, in states such as South Dakota, Indian children are "11 times more likely to be removed from their families and placed in foster care than non-Indian children." [21] Child custody loss classification is also high among decedents who are Hispanic. Although immigration status was not clearly documented within the structured NVDRS data, this may be related to well-documented increased incidents of child custody loss among Hispanic populations resulting from forced separations incurred while trying to navigate the US immigration system.[22] These findings may reflect trends in downstream parental mental health consequences of child custody loss and family separations known to be prevalent within these US populations.

Taken together, these classifiers may serve as valuable tools for monitoring trends in social isolation over time and across diverse demographic groups. These data and insights can be instrumental in developing more effective policies and interventions aimed at enhancing access to social support and addressing other specific needs of at-risk individuals and disproportionately affected populations. This could involve increasing access to mental health services, creating community support programs, or implementing policies that promote inclusivity and reduce disparities. Additionally, raising awareness about the health implications of social isolation can help empower individuals to seek out social connections and support.

By integrating these strategies, we can not only identify and address existing disparities but also proactively work towards reducing inequalities linked to social isolation. This approach underscores the importance of integrating data-driven insights into policy and intervention, ensuring a comprehensive and impactful response. The CDC's Suicide Prevention Resource for Action outlines comprehensive, evidence-based strategies for suicide prevention that communities and public health organizations can adopt to remedy existing health inequities, including social isolation and loneliness, and consequently reduce the risk of suicide and suicidal behaviors.[23]

### *Related Work*

While much has been learned from the NVDRS, researchers have largely used the structured variables; traditional qualitative methods are too labor intensive to use at scale. Despite the potential benefits to uncovering additional nuance and insights within LE/CME text narratives, these data are only beginning to be examined.[24]

Recent studies have applied similar NLP-based methods within the NVDRS for identifying intimate partner violence and assessing relevant predictors,[25,26] characterizing circumstantial antecedents for fire-arm related suicides among women,[27] and identifying social determinants of health such as economic, interpersonal, health and job-related problems.[28,29] NLP-related methodologies have also been used to improve overall classification of suicide and reduce potential data quality issues and racial biases in classification trends.[30,31] Our work stands with others who are working to leverage state of the art NLP methods to maximize the potential of LE/CME narratives to illuminate many aspects of violent death, from proximate correlates to nuanced context.

### *Limitations*

While this study offers several important contributions to the study of social isolation and suicide, it is important to consider key limitations. Classification frequencies observed in this study for chronic social isolation likely underrepresent actual incidence rates, given that the most socially isolated decedents are likely to lack evidence sources frequently referenced in the LE/CME narratives, which are often decedent next of kin, friends, spouses, or neighbors.

Classification models are likely influenced by many positive classes within most annotation samples. To create more generalizable models, it may be advantageous to annotate additional training data, or ensure that the training data are more balanced across positive and negative examples. Although we have taken every measure to ensure annotators agree in their labeling mechanism through collaborative development of coding ontologies, annotating the social isolation circumstances is still a challenging task due to its conceptually complex nature. The coding task may also be subject to annotator bias.[32] Additionally, while topic modeling on shorter-length summary fields was advantageous to initially identify more coherent topics, it may not fully capture the variability of language reflected in the full-length narratives. Lexicons developed in this study from summary fields may be able to be expanded in future iterations using word embeddings models trained on NVDRS narrative data.

Finally, it should be noted that although the NVDRS is one of the most comprehensive US data sources on violent deaths and suicide, many states changed or adopted data collection efforts only as of 2010, and thus we expect that there are differences in missingness based on decedent region before and after this time. Additionally, increases in prevalence may also reflect improvements in documentation and data collection.

Other studies have demonstrated that the prevalence of LE/CME narratives within NVDRS have increased over time as well.[12] Furthermore, our study period ends in 2020, and likely does not capture the full effect of the COVID-19 pandemic on increased social isolation, and potentially related suicides.

## MATERIALS AND METHODS

### *Experimental Design*

In this observational study we utilized a dataset from the National Violent Death Reporting System (NVDRS), which comprises 306,817 suicide incidents documented between 2002 and 2020, to apply NLP techniques to identify social-isolation related circumstances mentioned within suicide narratives which are not captured in current structured variables. The NVDRS consolidates comprehensive information on violent deaths from a variety of sources, including detailed narratives from law enforcement (LE) and coroner/medical examiner (CME) offices. Of the 306,817 decedents recorded, 216,237 (70.5%) of these had narrative information in at least one LE/CME free-text field.

### *Statistical Analysis*

Our NLP methodological framework is guided by Computational Grounded Theory, which emphasizes an iterative process of text data mining with the critical inclusion of human expertise.[33] Our approach combines advanced unsupervised and supervised learning techniques with human expert annotations to examine the circumstances surrounding suicides, specifically focusing on social isolation-related events and factors. The analytic process, as illustrated in Figure 2, includes: 1) topic modeling of suicide circumstance summaries guided by expert selection, 2) development of regular expression lexicons to search full LE/CME narratives, 3) annotation of these matches for topic relevance to train and evaluate supervised learning classifiers, and application of these classifiers across the entire NVDRS dataset.

### *1. BERTopic Topic Modeling*

Topic modeling began with the selection of cases from the NVDRS database that included text fields from LE/CME circumstance summaries, which occurred less frequently than full-length LE/CME narratives. LE circumstance summaries (n = 20,731, 6.8%) and CME circumstance summaries (n=30,584, 10.0%) provide shorter text summarizations (40-50 characters) of relevant contexts related to the suicides mentioned within the corresponding full LE/CME narratives. These circumstance summary texts were used for BERTopic topic modeling, an unsupervised machine learning approach that applies the BERT framework to derive meaningful topics from extensive textual data.[34] The shorter text within circumstance summaries resulted in less noise in topic model results than conducting the modeling on the full-length narratives. Summary topic model results could therefore be more easily understood and relevant topics identified by subject matter experts before identifying topics within the larger sample of full narrative texts. Further refining our BERTopic implementation, we conducted a grid search across 54 combinations of hyperparameters to optimize the topic clustering process prior to topic interpretation. This process is further detailed in the Supplemental Materials. The highest coherence model resulted in a total of 64 topics, including one uncategorized topic.

Next, the team conferred to select any topics identified in the topic model which were related to social isolation, conceptualized as either chronic social isolation, or events which could trigger intense social isolation. This method surfaced six social isolation-related topics: chronic social isolation, child custody loss, divorce/separation, breakup, loss of pet, and eviction/moving. These topics were used in the next step to identify key words and phrases which could be used to search through the full-length text narratives, which were longer, but far more prevalent than the summaries.

### 2. Regular Expression Lexicon Development and Matching

Following the unsupervised BERTopic modeling, we identified relevant key n-grams (from unigrams to trigrams) within the top 50 most frequent terms and phrases that were present in circumstance summaries categorized in each specific social isolation-related topic. Our team of annotators identified terms that were uniquely related to each topic under consideration. This step was essential to pinpoint terms related to social isolation or related events, which we subsequently used to conduct regex searches across the corpus of decedents' full-length text LE/CME narratives. At this point, we focused our analyses on the full-length narratives. To ensure the validity of these topics' association with suicide, an additional layer of expert annotation was performed on a sample of 100 full-length narratives matched with each of the six topics. This annotation process focused on confirming the relevance of the topics to the suicides reported in the NVDRS database. We conducted interrater agreement for the first 50 coded narratives for each narrative.

### 3. Supervised Learning Classifier Training, Evaluation, and Prediction

Each set of 100 annotated narratives were used to train a suite of supervised learning classifiers for each social isolation-related topic more precisely beyond keyword matching alone. These classifiers included Naive Bayes, Logistic Regression, Random Forest, and RoBERTa, with an 80/20 training/test split. Hyperparameter optimization was conducted across discrete values relevant to each model, described further in the Supplemental Materials. Finally, we evaluated classifiers, prioritizing high macro-F1 and positive-class recall rates, and saved those models which were used to predict a binary (relevant or non-relevant) classifier for each topic. After identifying the best performing models for each topic, we generated predictions for the 100 samples and conducted 1000 iterations of bootstrapping random slices of 80% of the dataset to generate confidence intervals for each performance metric.

## Acknowledgments

The National Violent Death Reporting System (NVDRS) is administered by the Centers for Disease Control and Prevention (CDC) by participating NVDRS states. The findings and conclusions of this study are those of the authors alone and do not necessarily represent the official position of the CDC or of participating NVDRS states.

**Funding:** Include all funding sources, including grant numbers, complete funding agency names, and recipient's initials. Each funding source should be listed in a separate paragraph such as:

National Institute on Drug Abuse R01DA057599 (AS, SD)

National Institute on Drug Abuse T32 DA0505552 (DW, SP)

**Author contributions:** Each author's contribution(s) to the paper should be listed (we suggest following the CRediT model with each CRediT role given its own line. No punctuation in the initials.
Conceptualization: DW, AS, SD
Methodology: DW, SR, SD, SP, AS
Investigation: DW, SR, SD, SP, AS
Visualization: DW
Supervision: AS, SS
Writing—original draft: DW, AS
Writing—review & editing: DW, SR, SD, SP, AS

**Competing interests:** Authors declare that they have no competing interests

**Data and materials availability:** Access to the NVDRS Restricted Access Database requires approval from the NVDRS RAD review committee, consisting of scientific and data analysis experts within CDC's National Center for Injury Prevention and Control. More information on application procedures can be found here: https://www.cdc.gov/nvdrs/about/nvdrs-data-access.html Code for analysis used in this study can be found at the following GitHub repository: https://github.com/drew-walkerr/nvdrs-social-isolation-classification.git

## Figures and Tables

You may include up to **a total of 10 figures and/or tables (combined)** throughout the manuscript. You should embed your figures within the Word file. A detailed description for figure preparation can be found on the topical style sheet for your field. For revised papers, include the captions at the end of the document and upload figures to CTS.

**Fig. 1: Social isolation-related suicide events over time, per 1000 suicides per year**

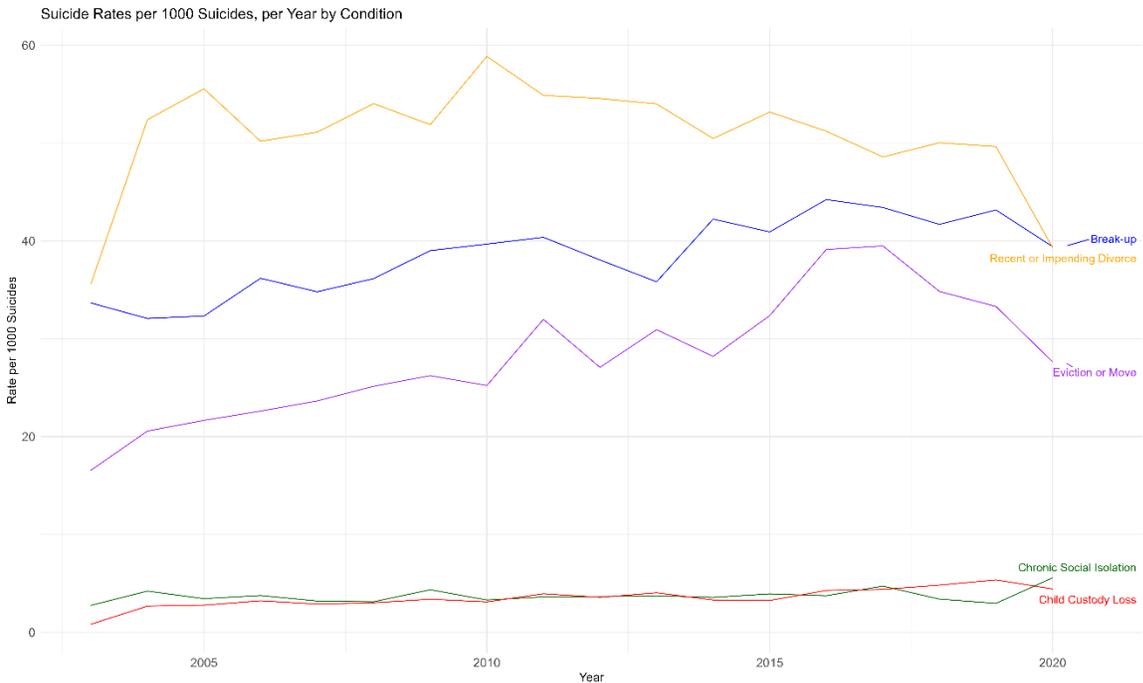

**Fig. 2:** Social isolation-related suicide circumstance identification analysis pipeline

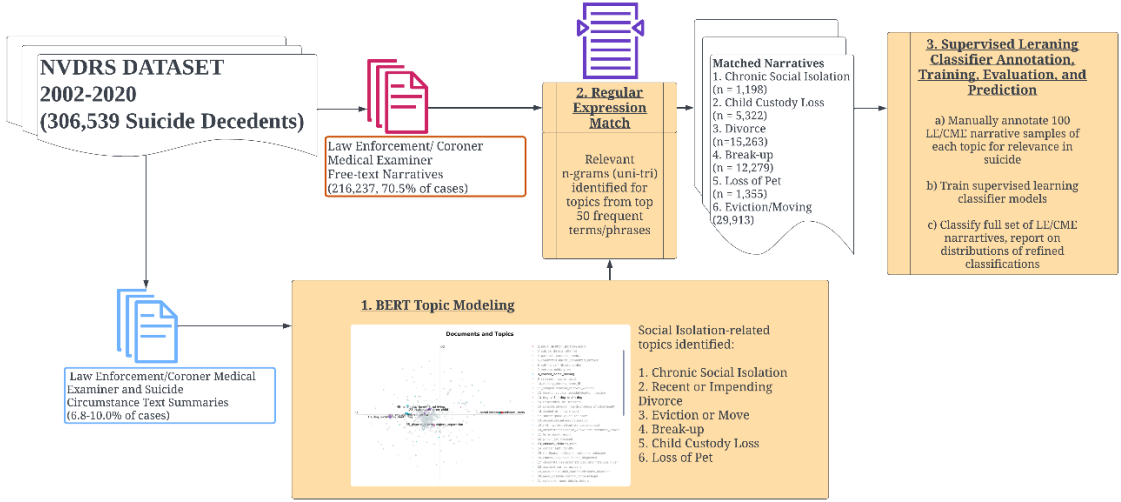

**Table 1 :** Demographics of suicide decedents in NVDRS, from 2002-2020.

|  | Frequency (%), or Mean (SD), Median[Min, Max] N = 306,817 decedents |
|---|---|
| **Sex** | |
| Female | 67,192 (21.9%) |
| Male | 239,616 (78.1%) |
| Unknown | 9 (.002%) |
| | |
| **Age (years)** | 46.3 (18.4) |
| | 46 [10, 106] |
| **Race/Ethnicity** | |
| American Indian/Alaska Native, non-Hispanic | 3928 (1.3%) |
| Asian/Pacific Islander, non-Hispanic | 7024 (2.3%) |
| Black or African American, non-Hispanic | 19,949 (6.5%) |
| Hispanic | 19,337 (6.3%) |
| Other/Unspecified, non-Hispanic | 894 (0.3%) |
| Two or more races, non-Hispanic | 3423 (1.1%) |
| White, non-Hispanic | 251,997 (82.1%) |
| Unknown | 265 (0.1%) |
| | |
| **Sexual Orientation** | |
| Bisexual | 217 (0.1%) |
| Gay | 1045 (0.3%) |
| Heterosexual | 30,819 (10.0%) |
| Lesbian | 514 (0.2%) |
| Unspecified sexual minority | 159 (0.1%) |
| Not explicitly reported | 274,063  (89.3%) |
| | |
| **Transgender** | |
| No, not available, unknown | 306,263 (99.8%) |
| Yes | 554 (0.2%) |
| | |
| **Marital Status** | 65,080 (21.2%) |
| Divorced | 99,590 (32.5%) |
| Married/Civil Union/Domestic Partnership | 7,534 (2.5%) |
| Married/Civil Union/Domestic Partnership, but separated | 109,189 (35.6%) |
| Never Married | 3,959 (1.3%) |
| Single, not otherwise specified | 17, 878 (5.8%) |
| Widowed | 3587 (1.2%) |
| Unknown | |
| | |
| **Relationship Status** | 82,535 (26.9%) |
| Currently in a relationship | 18,194 (5.9%) |
| Not currently in a relationship | 206,088 (67.2%) |
| Unknown | |
| | |
| **Homeless** | 287,594 (93.7%) |
| | 3,520 (1.1%) |

| No | 15,703 (5.12%) |
| Yes | |
| Unknown | |
| | 248,264 (80.9%) |
| **Physical Health Problem** | 58,553 (19.1%) |
| No, Not available, Unknown | |
| Yes | 3.61 (4.16) |
| | 2 [1, 132] |
| **Number of Substances detected by toxicology** | Missing: 173,886 (56.7%) |

**Table 2:** Regular expression matches and annotation results for social isolation-related narrative topics

| Social Isolation Life Event Topic | Total Narrative Regex Matches, % of total decedents, N = 306,817 | Sample (n=100) %Relevant | Interrater agreement for first 50 samples (Kappa) | Examples from Law Enforcement and Coroner/Medical Examiner Narratives |
|---|---|---|---|---|
| Social Isolation (Chronic) | 1,198 (0.39%) | 94% | 94% (k = .85, p = 1.38 x 10^-9) | "... victim's sister was contacted and she explained that the <u>victim was very much a loner</u> and was very cynical." |
| Impending or Recent Divorce | 15,263 (4.97%) | 66% | 72% (k = .24, p = .01) | "Victim and Victim's spouse <u>are in the process of getting a divorce</u>." |
| Eviction or Recent Move | 29,913 (9.75%) | 36% | 72% (k = .407, p = .004) | "...The day of the incident the <u>Victim was being evicted from their residence</u>." |
| Break-up | 12,279 (4.00%) | 100% | 98% (k = NA) | "...Family members stated that the Victim had been upset over the <u>recent break up</u> with his girlfriend." |
| Child Custody Loss | 5,322 (1.73%) | 50% | 90% (k = .80, p = 1.21 x 10^-8) | "...She [Victim] <u>lost custody of their children</u> due to domestic violence issues with their former spouse." |
| Loss of pet | 1,355 (0.44%) | 28% | 98% | "...He [Victim's brother] also said <u>the victim's dog</u> |

|  |  |  | (k = .905, p = 1.29 x 10^-10) | <u>died recently</u> and she was very upset about that. " |
|---|---|---|---|---|
| **Total narratives with at least one match** | **56,947 (18.56%)** |  |  |  |

**Table 3.** Bivariate logistic Regressions showing demographic predictors of social isolation-related topic narrative classifications. Odds ratios of classification (95%CI)

| | Chronic Social Isolation | Impending or Recent Divorce | Eviction | Break-up | Child Custody Loss |
|---|---|---|---|---|---|
| **Sex (Ref = Female)^** | **1.44 (1.24, 1.69)\*\*** | **1.31 (1.26, 1.37)\*\*** | **.88 (.84, .92)\*\*** | **1.13 (1.08, 1.18)\*\*** | **.57 (.51, .65)\*\*** |
| **Race/Ethnicity (Ref = White)** | | | | | |
| American Indian/Alaska Native | .62 (.31, 1.10) | **.48 (.39, .58)\*\*** | .86 (.71, 1.04) | **1.46 (1.27, 1.68)\*\*** | **3.02 (2.20, 4.04)\*\*** |
| Asian/Paific Islander | 1.16 (.80, 1.61) | **.74 (.66, .84)\*\*** | .91 (.79, 1.05) | .87 (.76, .99) | .61 (.36, .96) |
| Black or African American | **.62 (.46, .81)\*** | **.46 (.42, .50)\*\*** | .88 (.81, .96) | 1.02 (.95, 1.10) | 1.11 (.88, 1.38) |
| Hispanic | .80 (.61, 1.02) | **.81 (.75, .87)\*\*** | **1.18 (1.09, 1.28)\*\*** | **1.61 (1.51, 1.72)\*\*** | **1.68 (1.38, 2.02)\*\*** |
| Other/Unspecified | .83 (.20, 2.15) | **.51 (.34, .75)\*** | 1.13 (.77, 1.59) | 1.03 (.72, 1.42) | .90 (.22, 2.34) |
| Two or more races | **5.81 (.56, 1.64)** | 1.0  (.86, 1.16) | .98 (.80, 1.19) | **1.74 (1.52, 2.00)\*\*** | 1.72 (1.09, 2.57) |
| **Sexual Orientation (ref = Heterosexual)** | | | | | |
| Bisexual | 4.10 (1.00, 10.98) | .85 (.49, 1.37) | 1.23 (.63, 2.15) | 1.86 (1.18, 2.79) | .71 (.04, 3.18) |
| Gay | **3.68 (1.97, 6.33)\*\*** | **.28 (.19, .41)\*\*** | 1.18 (.88, 1.56) | **1.90 (1.55, 2.31)\*\*** | .74 (.26, 1.62) |
| Lesbian | .57 (.03, 2.56) | .59 (.39, .85) | .79 (.47, 1.24) | **3.26 (2.57, 4.08)\*\*** | .90 (.22, 2.38) |
| Unspecified sexual minority | 3.72 (.61, 11.87) | .42 (.16, .86) | .29 (.05, .92) | 1.91 (1.13, 3.04) | 0 (0,0) |
| **Transgender (ref = cisgender)** | 3.29 (1.41, 6.42) | **.24 (.10, .48)\*\*** | **1.11** (.59, 1.57) | 1.04 (.67, 1.54) | 0 (0,0) |
| **Marital Status (Ref: Married/Civil Union/Domestic Partnership)** | | | | | |
| Divorced | **3.34 (2.68, 4.19)\*\*** | **.68 (.65, .71)\*\*** | **1.33 (1.25, 1.40)\*\*** | **6.42 (5.89, 7.02)\*\*** | **2.06 (1.75, 2.42)\*\*** |
| Married/Civil Union/Domestic Partnership, but separated | **3.43 (2.24, 5.09)\*\*** | **6.61 (6.27, 6.96)\*\*** | **2.34 (2.02, 2.48)\*\*** | **4.96 (4.25, 5.77)\*\*** | **2.65 (1.94, 3.53)\*\*** |
| Never Married | **5.81 (4.78, 7.12)\*\*** | **.07 (.02, .07)\*\*** | **1.11 (1.05, 1.17)\*\*** | **12.89 (11.89, 14)\*\*** | **1.75 (1.51, 2.03)\*\*** |
| Single, not otherwise specified | 2.03 (.95, 3.77) | **.04 (.02, .07)\*\*** | .87 (.71, 1.07) | **15.32 (13.38, 17.51)\*\*** | 1.55 (.86, 2.48) |
| | **3.04 (2.21, 4.14)\*\*** | **.10 (.08, .12)\*\*** | 1.11 (1.01, 1.22) | **1.35 (1.13, 1.61)\*** | .81 (.57, 1.12) |

Widowed

| | | | | | |
|---|---|---|---|---|---|
| **Relationship Status (Ref: Currently in a relationship)** Not currently in a relationship | **6.97 (5.61, 8.69)\*\*** | **1.79 (1.69, 1.89)\*\*** | **1.58 (1.47, 1.70)\*\*** | **11.42 (10.79, 12.08)\*\*** | 1.36 (1.10, 1.68) |
| | .84 (.45, 1.42) | | | | |
| **Homeless** | 1.04 (.90, 1.20) | **.73 (.61, .87)\*\*** | **1.42 (1.20, 1.66)\*\*** | .96 (.81, 1.14) | **2.08 (1.41, 2.94)\*\*** |
| **Physical Health Problem** | 1.00 (.98, 1.02) | **.54 (.51, .57)\*\*** | 1.05 (1.00, 1.11) | **.27 (.25, .29)\*\*** | **.51 (.43, .61)\*\*** |
| **Number of Substances** | | .99 (.99, 1.00) | **1.02 (1.01, 1.02)\*\*** | .99 (.99, 1.00) | **1.03 (1.01, 1.04)\*\*** |

\*p is significant at <.00167 value (Bonferroni correction for 30 tests)
\*\*p is significant at <.0001 value
^ Sex was derived from either death certificates, coroner/medical examiner, or law enforcement

**Supplementary Materials**

*Topic modeling optimization*

      Our application of the BERTopic model involved customizing several components to accommodate our specific research needs. For text vectorization, we utilized `CountVectorizer()`, including both unigrams and bigrams, to capture both words and short phrases. To enhance topic discrimination, we integrated `TfidfTransformer()` to apply TF-IDF weights to each token within the documents, which helped us identify unique words across the various suicide circumstance summaries.

      Hyperparameters adjusted for topic modeling included the minimum cluster size, determined by the `HDBSCAN()` function from Scikit-Learn, which can identify clusters of varying densities.[34] We also optimized the `n_components` and `min_dist` settings of the `UMAP()` function, which facilitated non-linear dimensionality reduction and ensured appropriate spacing between clusters.[35] Model performance was then evaluated based on coherence scores calculated using the `get_coherence()` function from the Gensim package, providing a quantitative measure of the model's ability to produce interpretable topics.[36,37] Of all models ran, our highest coherence (.946) was found with our model with the following hyperparameters: 57 minimum distance, 15 neighbors, 3 components, .01 min_dist. This resulted in a total of 64 topics, including one uncategorized topic.

*Supervised model optimization*

24      Hyperparameter optimization assessed best scores for models for logistic regression across regularization parameter (C ) values of : 0.01,

25      0.1, 1.0 ; random forest hyperparameter values included n_estimators (number of trees) of 100, 200, and 300, max depth of trees of none, 10, 20,

26      and minimum samples split: 2, 5, 10; naive bayes values included alpha scores of .1, .5, and 1.0; RoBERTa values tested included max token length

27      of 128, 512, batch size of 16, across 10 epochs, with learning rates of $5\text{x}10^{-6}$ and $1\text{x}10^{-5}$.

28

29

**Supplemental Table 1: Social Isolation Event NVDRS Narrative Classifier Performance (100 samples)**

| Bias Feature | Model | Accuracy | Precision (Positive) | Recall (Positive) | F1 (Positive) | Macro Precision | Macro Recall | Macro F1 |
|---|---|---|---|---|---|---|---|---|
| **Social Isolation** | RoBERTA | .88 (.81, .93) | .88 (.81, .93) | 1 | .94 (.90, .97) | .44 ( .41, .48). | .50 (.50, .50) | .47 (.45, .48) |
| | **Logistic Regression** | **.86 (.71,1)** | **.86 (.71,1)** | **1** | **.92 (.83,1)** | **.86 (.71,1)** | **1** | **.92 (.83,1)** |
| | Naive Bayes | .86 (.71,1) | .86 (.71,1) | 1 | .92 (.83,1) | .86 (.71,1) | 1 | .92 (.83,1) |
| | Random Forest | .86 (.71,1) | .86 (.71,1) | 1 | .92 (.83,1) | .86 (.71,1) | 1 | .92 (.83,1) |
| **Recent or Impending Divorce** | RoBERTA | .76 (.67, .82) | .76 (.68, .83) | 1 | .86 (.81, .91) | .38 (.34, .41) | .50 (.50, .50) | .43 (.41, .45) |
| | **Logistic Regression** | **.66 (.66, .66)** | **.75 (.58, .92)** | **.75 (.58, .92)** | **1** | **.85 (.74, .96)** | **.75 (.58, .92)** | **1** |
| | Naive Bayes | .66 (.66, .66) | .75 (.58, .92) | .75 (.58, .92) | 1 | .85 (.74, .96) | .75 (.58, .92) | 1 |
| | Random Forest | .66 (.66, .66) | .75 (.58, .92) | .75 (.58, .92) | 1 | .85 (.74, .96) | .75 (.58, .92) | 1 |
| **Eviction or Move** | RoBERTA | .68 (.60, .77) | .68 (.53, .82) | .53 (.39, .67) | .59 (.47, .71) | .68 (.59, .77) | .67 (.58, .75) | .67 (.57, .75) |
| | Logistic Regression | .65 (.43, .83) | .66 (.2, 1) | .40 (.1, .70) | .48 (.14, .75) | .66 (.2, 1) | .40 (.1, .70) | .48 (.14, .75) |
| | **Naive Bayes** | **.70 (.52,.87)** | **.79 (.25, 1)** | **.40 (.1,.75)** | **.52 (.15, .80)** | **.79 (.25, 1)** | **.40 (.1, .75)** | **.52 (.15, .80)** |
| | Random Forest | .52 (.35, .74) | 0 | 0 | 0 | 0 | 0 | 0 |

| | | | | | | | |
|---|---|---|---|---|---|---|---|
| **Break Up** | RoBERTA | .97 (.95, 1) | .97 (.94, 1) | 1 (1,1) | .99 (.97, 1) | .49 (.47, 1) | .50 (.50, 1) | .49 (.48, 1) |
| | Logistic Regression | **1** | **.96 (.86, 1)** | **1** | **.98 (.93, 1)** | **.96 (.86, 1)** | **1** | **.98 (.93, 1)** |
| | Naive Bayes | 1 | .96 (.86, 1) | 1 | .98 (.93, 1) | .96 (.86, 1) | 1 | .98 (.93, 1) |
| | Random Forest | 1 | .96 (.86, 1) | 1 | .98 (.93, 1) | .96 (.86, 1) | 1 | .98 (.93, 1) |
| **Child Custody Loss** | RoBERTA | .46 (.36, .54) | .67 (0, 1) | .03 (0, .08) | .06 (0, .14) | .56 (.21, .76) | .51 (.48, .53) | .34 (.28, .41) |
| | **Logistic Regression** | **.87 (.74, 1)** | **1 (1, 1)** | **.78 (.5, 1)** | **.87 (.67, 1)** | **1 (1,1)** | **.78 (.5 , 1)** | **.87 (.67, 1)** |
| | Naive Bayes | .78 (.61, .91) | .90 (.67, 1) | .69 (.42, .93) | .77 (.56, .94) | .90 (.67, 1) | .69 (.42, .93) | .77 (.56, .94) |
| | Random Forest | .61 (.39, .78) | .63 (.38, .85) | .77 (.5, 1) | .68 (.45, .86) | .63 (.38, .85) | .77 (.5, 1) | .68 (.45, .86) |
| **Pet loss** | RoBERTA | .70 (.61, .77) | 0 (0,0) | 0 (0,0) | 0 (0,0) | .35 (.61, .39) | .50 (.50, .50) | .41 (.38, .44) |
| | **Logistic Regression** | **.63 (.63, .63)** | **.78 (.61, .91)** | **.87 (0,1)** | **.28 (0, .67)** | **.41 (0, .80)** | **.87 (0, 1)** | **.28 (0, .67)** |
| | Naive Bayes | .73 (.73, .73) | .69 (.52, .87) | 0 | 0 | 0 | 0 | 0 |
| | Random Forest | .73 (.73, .73) | .69 (.52, .87) | 0 | 0 | 0 | 0 | 0 |

31
32
33
34
35

36 **Supplemental Table 2: Total rate of narrative classification predictions of each social isolation related topic across NVDRS law**
37 **enforcement and coroner medical examiner narratives, percent predicted positive, and normalized total rate per 1000 suicides**
38

| Topic | Regex Matches, N | Refined Supervised Learning Predictions, N | Percentage Predicted Positive | Total rate per 1000 suicides |
|---|---|---|---|---|
| Chronic Social Isolation | 1198 | 1198 | 1 | 3.905 |
| Recent or impending divorce | 15331 | 15311 | 0.998 | 49.977 |
| Recent eviction/move | 29977 | 9468 | 0.316 | 30.866 |
| Recent breakup | 12311 | 12311 | 1 | 40.126 |
| Child Custody Loss | 5326 | 1231 | 0.231 | 4.012 |

39